\documentclass[10pt,twocolumn,letterpaper]{article}

\usepackage{btas}
\usepackage{times}
\usepackage{graphicx}
\usepackage{amsmath}   
\usepackage{amssymb}
\usepackage{epstopdf}
\graphicspath{ {images/} }
\usepackage{url}
\usepackage{subfigure}
\usepackage{enumitem}

\btasfinalcopy 

\ifbtasfinal\fi

\begin{document}

\title{Active User Authentication for Smartphones: A Challenge\\ Data Set and Benchmark Results}

\author{Upal Mahbub$^{1}$ \quad Sayantan Sarkar$^{1}$ \quad Vishal M. Patel$^{2}$  \quad Rama Chellappa$^{1}$\\
$^{1}$Department of Electrical and Computer Engineering and the Center for Automation Research, \\UMIACS, University of Maryland, College Park, MD 20742\\
{\tt\small \{umahbub, ssarkar2, rama\}@umiacs.umd.edu}\\$^{2}$Rutgers, The State University of New Jersey, 508 CoRE, 94 Brett Rd, Piscataway, NJ 08854\\
{\tt\small vishal.m.patel@rutgers.edu}\thanks{This work was done in partnership with and supported by Google Advanced Technology and Projects (ATAP), a Skunk Works-inspired team chartered to deliver breakthrough innovations with end-to-end product development based on cutting edge research and a cooperative agreement FA8750-13-2-0279 from DARPA.} }

\vskip -5pt

\maketitle
\thispagestyle{empty}

\begin{abstract}
In this paper, automated user verification techniques for smartphones are investigated. A unique non-commercial dataset, the University of Maryland Active Authentication Dataset 02 (UMDAA-02)  for multi-modal user authentication research is introduced. This paper focuses on three sensors - front camera, touch sensor and location service while providing a general description for other modalities. Benchmark results for face detection, face verification, touch-based user identification and location-based next-place prediction are presented, which indicate that more robust methods fine-tuned to the mobile platform are needed to achieve satisfactory verification accuracy. The dataset will be made available to the research community for promoting additional research.
\end{abstract}

\section{Introduction}
The recent proliferation of mobile devices like smartphones and tablets has given rise to security concerns about personal information stored in them. Studies show that users are more concerned about the security of their cell phones over laptops \cite{PhoneSecurity_Chin}. Though over $40\%$ of users in major U.S. cities have lost their phones or have been victims of phone theft  \cite{SmartPhoneNotSmartEnough:Fischer}, industry surveys estimate that  $34\%$ of smartphone users in the U.S. do not lock their phones with passwords \cite{ConsumerReport:PhoneSecurity}. This contradictory behavior is due to the time-consuming, cumbersome and error-prone hassles of entering passwords on virtual keyboards or due to users' beliefs that extra passwords are not needed \cite{SmartPhoneNotSmartEnough:Fischer}.  $76\%$ attacks on smart phones exploit weak passwords \cite{VerizonReportDBIR}, but users still prefer those over stronger passwords, as the stronger passwords are difficult to remember and type, especially since the average cell phone user checks their smartphone device $150$ times per day \cite{MeekerReport2013}.

\begin{figure}[t]
\centering
\includegraphics[width=0.45\textwidth]{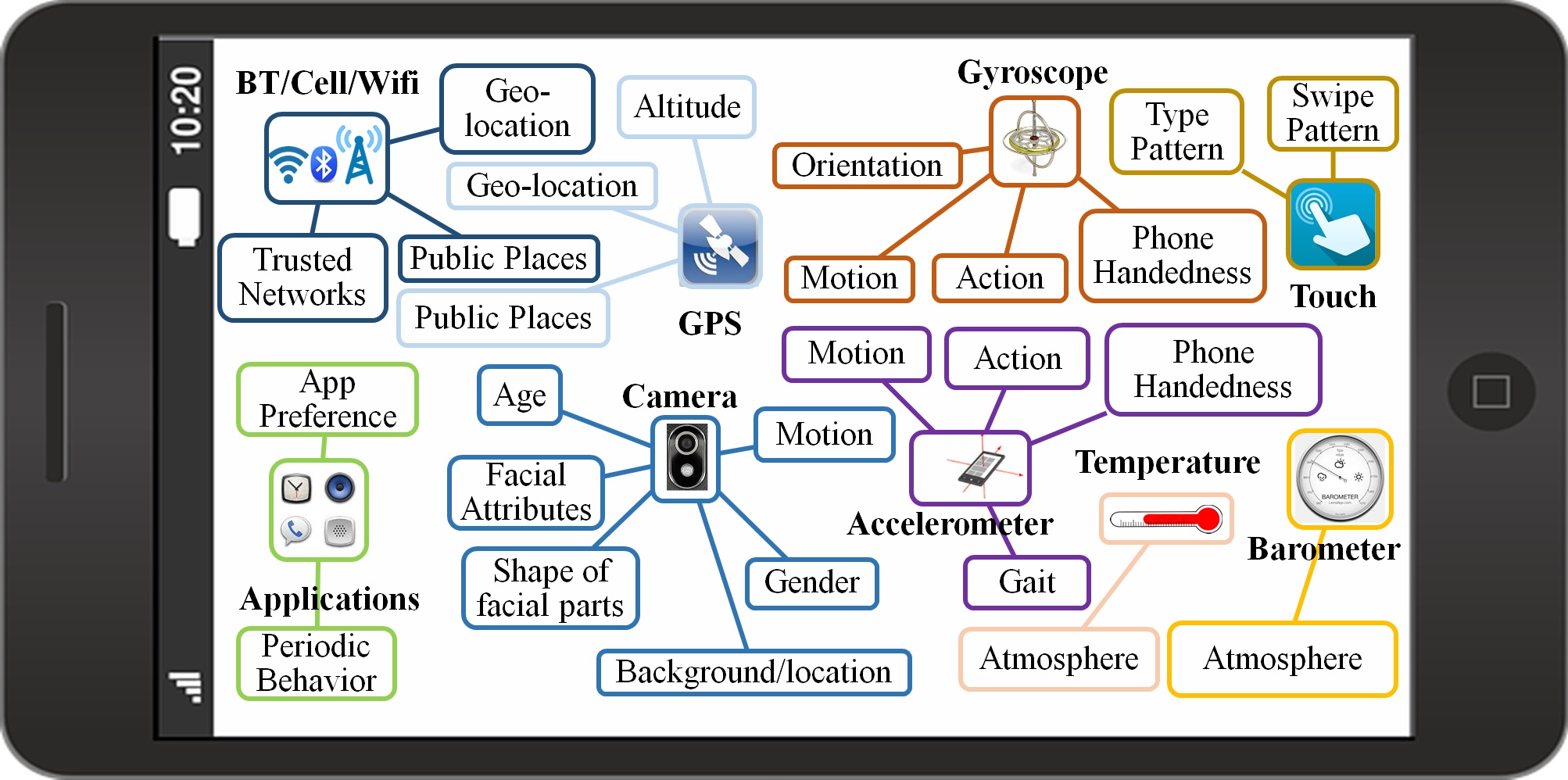}
\vskip -0pt
\caption{Association of smartphone sensors with behavioral and biometric information.}
\label{SensorInfo}
\vskip -15pt
\end{figure}

Going beyond traditional passwords and fingerprint-based one-time authentication, the concept of Active Authentication (AA) has emerged recently \cite{VMP_SPM_AA_2016}, where the enrolled user is authenticated continuously in the background based on the user's biometrics such as front camera face capture \cite{AA_Samangouei}, \cite{AA_Fathy}, touch screen gesture \cite{TouchAA_Feng}, \cite{umd_Dataset}, typing pattern \cite{TypingAA_Arujo} etc. Conceptually, in an AA system users do not password-lock the phone at all. When a user uses the phone, the AA system compares the usage pattern with the enrolled user's pattern of use. If the system deems that the usage patterns are sufficiently similar, the phone's full functionality (including sensitive applications and data) is made available, else  it blocks the current user. At present, most of the AA systems are based on face, touch and typing pattern biometrics. As shown in Fig. \ref{SensorInfo}, modern smartphones provide multiple sensors associated with a variety of behavioral and physiological biometric information, however research on multi-modal authentication using multi-sensor data is lagging behind, because of paucity of datasets.
 
The first non-commercial dataset on smartphone usage containing a wide range of sensor data, namely the University of Maryland Active Authentication Dataset 02 (UMDAA-02), is introduced in this paper. Unlike task-based data collection schemes, the data collection was passive and hence is representative of the natural, regular smartphone usage by the volunteers. The data collection application ran on the Nexus-5 device, completely in the background, saving sensor data and periodically uploading the data to a secure online location.

The benchmark results of $4$ experiments on the UMDAA-02 dataset are reported in this paper. Face is the most widely used biometric, but the images captured by the front-facing camera of smartphones present certain challenges such as partial face detection under occlusion and large variations in pose and illumination. The face images in the UMDAA-02 dataset are difficult to detect and the performances of traditional face detection methods are explored on a smaller annotated subset of the dataset. Next, faces of the annotated subset are verified using multiple state-of-the-art features and distance measures. On the full dataset, swipe-based user identification has been performed. Also, utilizing the user's geolocation information, the next place prediction experiment is performed which can be useful in AA research when fusing multiple modalities.

\section{Previous Works}
Among the AA techniques, the most explored are based on faces \cite{AA_Fathy}, \cite{AA_Samangouei}, touch/swipe signature\cite{Touch_SerwaddaPW13}, multi-modal fusion \cite{umd_Dataset}, gait \cite{AA_Gait} and device movement-patterns/accelerometer  \cite{AccDataAuth_Primo2014}, \cite{ContAuth_AJain}. Face-based authentication, though most accurate, requires more computational power and can cause faster battery drain if the images are  captured frequently. On the other hand, swipe and accelerometer data alone are not discriminative enough. Among the other AA approaches, in \cite{AA_StylometryAppWeb_Friedman}, the authors fused stylometry with application usage, web browsing data and location information.

Various protocols for AA with and without multi-modal fusion have been suggested over the years. In \cite{ProgressiveAuth_Riva2012}, the authors explored the idea of progressive or risk-based authentication by combining multiple verification signals to determine the user’s level of authenticity. The AA system surfaces only when this level is too low for the content being requested. In \cite{ContextAware_Grady}, the authors proposed context aware protocols for more flexible yet robust authentication. In \cite{Fusion_Ross}, the authors discuss three possible levels of fusion (a) fusion at feature level, (b) fusion at score level, and (c) fusion at decision level. Different fusion algorithms based on k-Nearest Neighbour classifiers, Support Vector Machines, decision trees, Bayesian methods, Gaussian Mixture Models (GMM) have been employed. \cite{Fusion_Ross}, \cite{Fusion_Damousis}.

The MOBIO dataset \cite{Mobio_2012} is a well-known dataset for  face-based AA research. It contains $61$ hours of audio-visual data from a NOKIA N93i phone (and a 2008 Mac-book laptop) with $12$ distinct sessions of $150$ participants spread over several weeks. However, since users were required to  position their head inside a certain elliptical box  within the scene while capturing the data, the face images of this dataset do not represent real-life acquisition scenarios.

Faces captured by the front camera (and also screen touch data) of University of Maryland Active Authentication Dataset (UMDAA-01) \cite{umd_Dataset}\cite{AA_Samangouei} of $50$ users are unconstrained and hence presents a more realistic and challenging scenario for face-based continuous authentication where partially visible, frontal and non-frontal faces under various illumination conditions are available. In \cite{FSFD_Mahbub} and \cite{Sarkar_DeepFeatureFD}, the authors introduced facial segment-based face detection (FSFD) method and deep feature-based face detection for UMDAA-01-FD which is a small annotated subset of the UMDAA-01 dataset, respectively, and showed that the partial face detection capabilities of these methods make them suitable candidates for mobile front-camera face detection.

The MIT Reality Dataset \cite{MITRealityDataset} consists of call logs, Bluetooth devices in proximity, cell tower IDs, application usage, and phone status (such as charging and idle) information from $100$ Nokia-6600 smart phones users collected over 450,000 hours. Since it focused on analyzing social behavior of the subjects, it does not contain vital biometrics such as face and touch. The  Rice  Livelab  dataset  \cite{Rice_livelab_dataset} consists of information on application  usage,  wifi  networks,  cell  towers,  GPS  readings, battery  usage  and  accelerometer  output of 35 users, collected from iPhone 3GS devices over durations ranging from a few days to less than a year.

The largest known dataset on smartphone usage is the Google's Project Abacus data set consisting of $27.62$ TB of smartphone sensor signals collected from approximately $1500$ users for six months on Nexus $5$ phones \cite{NataliaNeverova_Abacus}. Data was collected for the front-facing camera, touchscreen and keyboard, gyroscope, accelerometer, magnetometer, ambient light sensor, GPS, Bluetooth, WiFi, cell antennae, app usage and on time statistics. Google also collected the $114$GB Project Move data set, which consists of smartphone inertial signals collected from $80$ volunteers over two months on $LG3$, $Nexus 5$, and $Nexus 6$ phones. The data collection was passive for both projects. To date, neither of these two datasets are available for the research community.

\section{Description of the UMDAA-02 Dataset}
\begin{table}
\small
\centering
\caption{Significant Information for Each Modality Per Session}
\begin{tabular}{p{2.2cm} p{5.3cm}}
\hline
Modality            & Information\\
\hline
\hline
Accelerometer       & Event Time, X, Y, Z\\
\hline
Gyroscope           & Event Time, X, Y, Z \\
\hline
Image               & Shutter Time, Filename\\
\hline
Bluetooth           & Developer, Paired/Unpaired Flag\\
\hline
Location            & Event Time, Lat., Long., Accuracy\\
\hline
Usage               & Event Time, $\%$ CPU, $\%$ Memory\\
\hline
Magnetic Field      & Event Time, X, Y, Z\\
\hline
Gravity             & Event Time, X, Y, Z\\
\hline
Connectivity       & Capture Time, Flag (Bluetooth, Gps, Wifi, Cell Network), Network Name and Code\\
\hline
Foreground  App Info            & Start Time, Duration, End Time, App Name, Launched From Home Flag\\
\hline
WiFi                & SSID, BSSID, Authentication Type, IP Address, RSSI\\
\hline
Ambient Light       & Event Time, Value\\
\hline
Ambient Cells       & MCC, CI, MNC, Sig. Strength, TAC\\
\hline
Screen              & Event Time, Key\\
\hline
Motion/Touch              & Event Time, Type, Pressure, Major-Minor Axis, Position\\
\hline
Call                & Event Time, Key\\
\hline
Key                 & Event Time, Pressure, Type, Key Code\\
\hline
Screen Res          & Event Time, X, Y\\
\hline
\end{tabular}
\label{GenInfoOnModalities}
\vskip -5pt
\end{table}

The UMDAA-02 data set consists of $141.14$ GB of smartphone sensor signals collected from $48$ volunteers on Nexus 5 phones over a period of 2 months (15 Oct. 2015 to 20 Dec. 2015). The data collection sensors include the front-facing camera, touchscreen, gyroscope, accelerometer, magnetometer, light sensor, GPS, Bluetooth, WiFi, proximity sensor, temperature sensor and pressure sensor. The data collection application also stored the timing of screen lock and unlock events, start and end time stamps of calls, currently running foreground application etc. The volunteers used the research phone as their primary device for a week and were given the option to stop data collection at will and review the stored data prior to sharing. 

In Table \ref{GenInfoOnModalities}, the most significant information for each modality associated with the sensor data is presented. Data for most of the modalities are stored when there is significant change in that modality. For example, the GPS data is stored at a rate proportional to the movement speed of the phone. The front camera images are captured only for the first $60$ seconds for each session at a rate of $3$ fps.

\begin{table}
\small
\centering
\caption{Information on UMDAA-02 and  UMDAA-02-FD Dataset}
\begin{tabular}{p{3.7cm} c c}
\hline
Description				& {\footnotesize UMDAA-02}			&{\footnotesize UMDAA-02-FD}\\
\hline
\hline
No. of Subjects	               &  $36$M, $12$F       &   $34$M, $10$F\\
\hline
Age Range (years)                       &$22-31$     &$22-31$\\
\hline
Avg. Days/User (days)       &$\backsim$ $10$         &$\backsim$ $10$ \\
\hline
Avg. Sessions/User    &$\backsim$ $248$     & $\backsim200$\\
\hline
Total Number of Images 	&$600712$   &$33209$ \\
\hline
No. of Images without Faces 	&$-$  	&$9060$\\ 
\hline
Avg. Images/User      &$\backsim$ $12515$      & $\backsim755$\\
\hline
Avg. Images/Session   &$\backsim51$   &$\backsim$ $4$\\
\hline
Min. no. of Image for a User		& $1038$ & $64$ \\
\hline
Max. no. of Image for a User	& $49023$	& $2787$ \\
\hline
\end{tabular}
\label{GenInfo}
\vskip -10pt
\end{table}

Some general information on the dataset is provided in Table \ref{GenInfo}. The usage information is arranged in `Sessions' which starts when the user unlocks the phone and ends when the phone goes to the locked state. The data is stored in nested folders with the year, month, day and start time of the session embedded in the folder names.

\section{Face Detection and User Verification}
\begin{figure}
\centering
\includegraphics[width=0.35\textwidth]{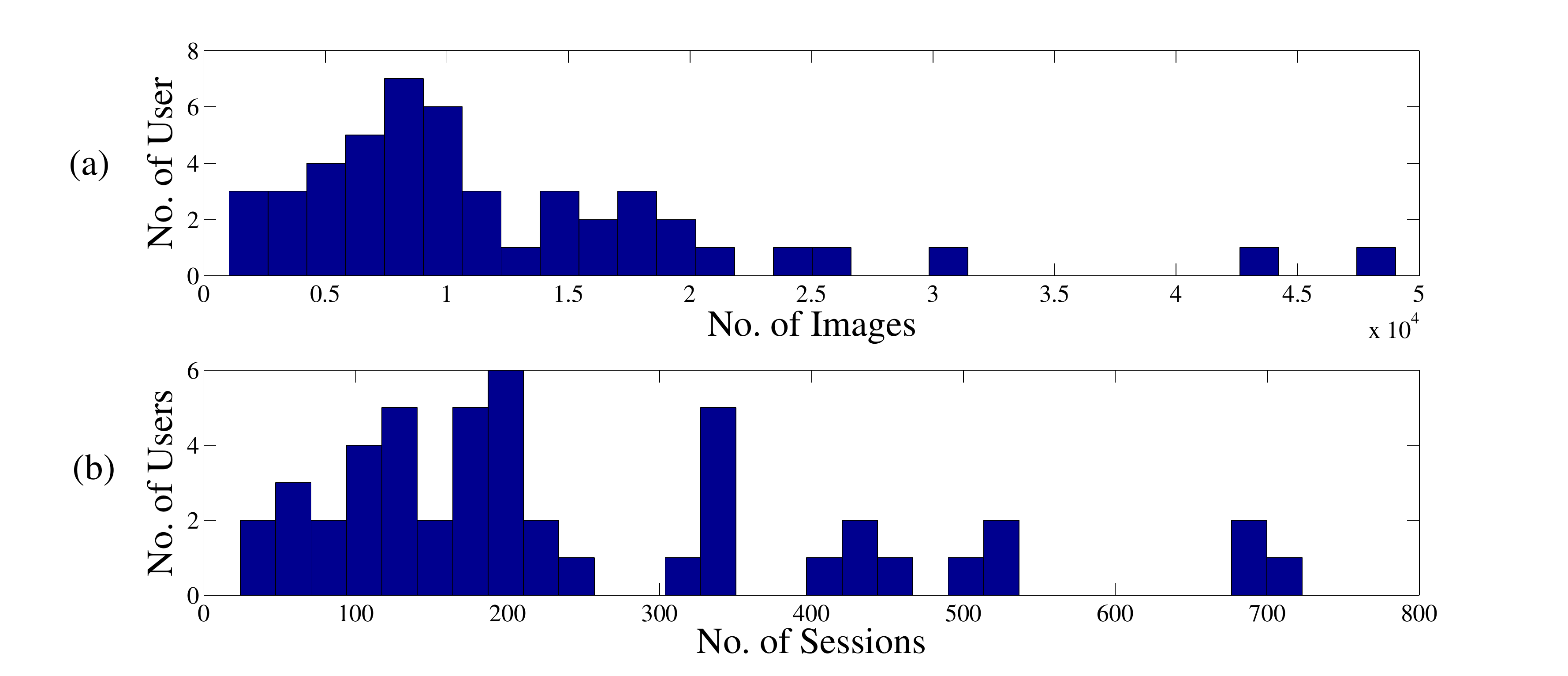}
\vskip -0pt
\caption{(a) Histogram of number of images per user, and (b) histogram of number of sessions per user.}
\label{ImgHist}
\vskip -5pt
\end{figure}

In this section, we describe face detection and verification tasks from faces captured by the front-facing camera. Fig.~\ref{ImgHist} shows histograms of number of images per user and the number of sessions per user. The number of images varies between $2000$ to $50,000$ per user and the number of sessions varies between $25$ and $750$, thus providing a large number of images for each user and session. 


\begin{figure}[t]
\centering
\includegraphics[width=0.35\textwidth]{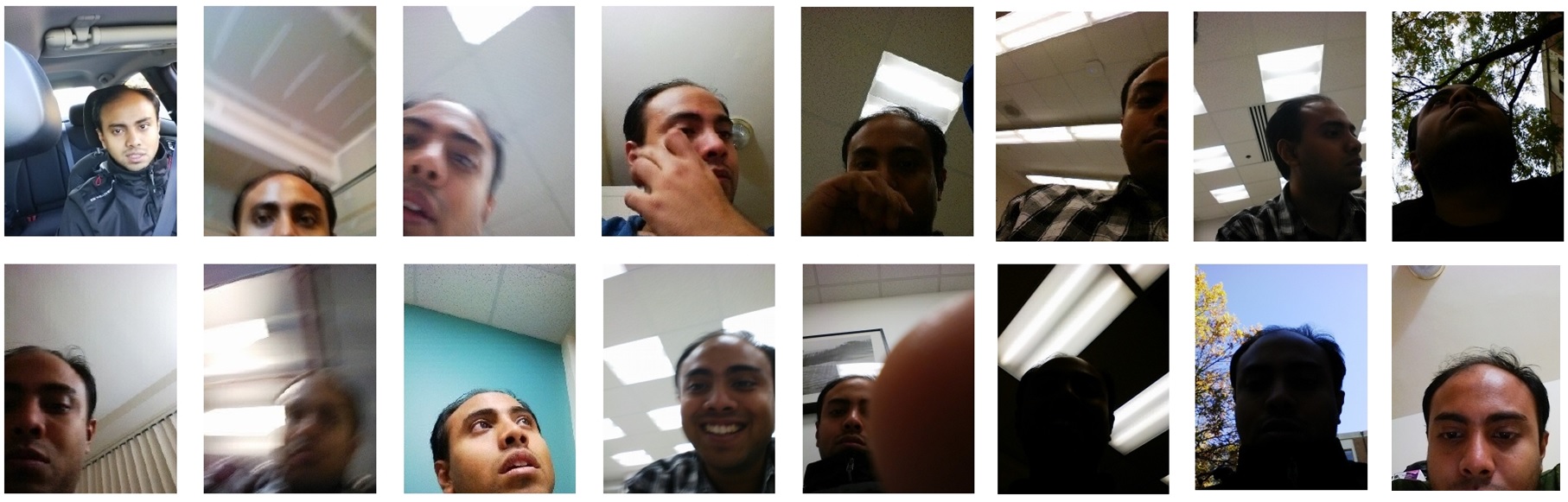}
\vskip 0pt
\caption{Sample images from one of the users showing a wide variety of pose, illumination, occlusion and expression variations.}
\label{SampleImages}
\vskip -15pt
\end{figure}

\noindent {\bf{UMDAA-02-FD Face Detection Dataset:}}
State-of-the-art face detection algorithms that perform satisfactorily on datasets like faces-in-the-wild \cite{LFWTech}, \cite{AFLWDataset} are not suitable for detecting partially visible faces that are typically present in the UMDAA-02 dataset. Moreover, for practical implementation purposes, the algorithm must be very fast and have a high recall rate to ensure continuous authentication \cite{FSFD_Mahbub}. A few sample images are shown in Fig. \ref{SampleImages} which shows that the faces suffer from partial visibility, illumination changes, occlusion and wide variation in poses and facial expressions. 

\begin{figure}[t]
\centering
\includegraphics[width=0.3\textwidth]{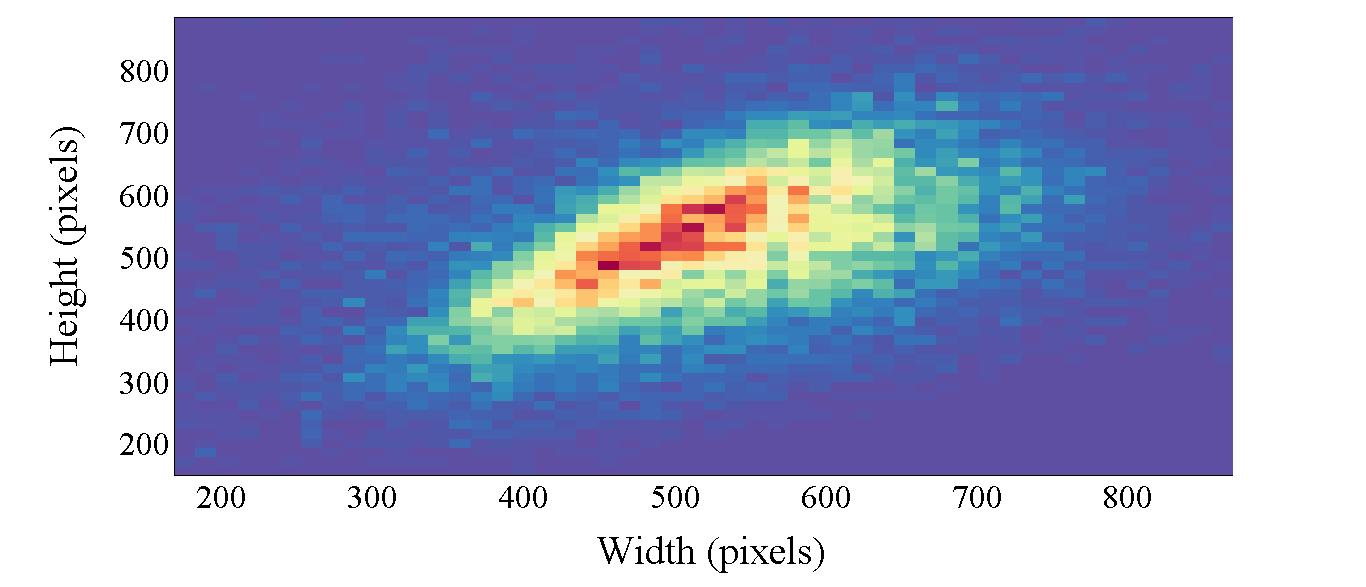}
\vskip -2pt
\caption{Distribution of bounding box width and heights}
\label{HWHist}
\vskip -5pt
\end{figure}

Excluding the data of $5$ users from a phone whose front camera malfunctioned during data collection phases, a set of $33209$ images was selected from all  sessions of the remaining $43$ users at an interval of $7$ seconds. The images were manually annotated for ground truth face bounding box, face orientation and five landmarks - left eye, right eye, nose, left and right corners of the mouth to create the UMDAA-02 face detection dataset (UMDAA-02-FD). Some information on the UMDAA-02-FD is provided in Table \ref{GenInfo}. The chronology and session information of all the images are also available.  The histogram of face height and width distribution shown in  Fig. \ref{HWHist} indicates that face widths vary approximately from $400$ to $650$ pixels, while face heights vary approximately from  $300$ to $700$ pixels. The database contains many partial faces as can be seen from the extremities of the distribution, information from which can help tune the hyper-parameters of face detectors.

\noindent {\bf{Evaluation of Face Detection Performances:}}
Accuracy and F1-score measures are adopted as evaluation metrics for face detection to ensure that both precision and recall performances are taken into consideration. The processing time per image is also measured to analyze the suitability for real-time operations. Prior to face detection, the images are down sampled by $4$ to ensure reasonable processing time for all algorithms. $50\%$ intersection-over-union overlap between the detection results and the ground truth bounding box is considered to be the threshold for correct detection.

\begin{table}[t]
\centering
\caption{Comparison between FD methods at $50\%$ overlap}
\vskip 0pt
\begin{tabular}{p{2.0cm} c c p{1.8cm}}
\hline
Method	& Accuracy 	& F1-Score		& Time/Image(s)\\
\hline
\hline
VJ \cite{VJFull}	& 60.24 		&64.50 	& \textbf{0.16}	\\
\hline
DPM	\cite{Ramanan:2012:FDP:2354409.2355119}	& 62.62		& 65.50		& 5.51	\\
\hline
LAEO \cite{LAEOdataset} 	& 19.40   	& 32.49 	& 4.57	\\
\hline
FSFD($C_{best}$)\cite{FSFD_Mahbub} & 73.48		& 79.11 	& 0.68\\
\hline
DP2MFD\cite{RRanjan_DeepPyramidFD} & \textbf{76.15}		& \textbf{82.83} 	& 15.0(CPU), 0.8(GPU)\\
\hline
\end{tabular}
\label{FD_Res}
\vskip -10pt
\end{table}

The performances of four face detection algorithms on the UMDAA-02-FD dataset are presented in Table \ref{FD_Res}. The recently proposed Facial Segment-Based Face Detector (FSFD) algorithm \cite{FSFD_Mahbub} (with number of random subset $\zeta=20$ and minimum number of segments $c=2$), which is specifically designed for detecting partial faces, performs better than other popular non-commercial detectors like Viola-Jones (VJ) \cite{VJFull} and Deformable Part-based Model (DPM) \cite{Ramanan:2012:FDP:2354409.2355119} and in reasonable processing time. Another recent FD technique, the Deep Pyramid Deformable Part Model (DP2MFD) \cite{RRanjan_DeepPyramidFD} utilizes normalized convolutional neural network (CNN) features. It outperforms all the other methods in terms of Accuracy and F1-Score but the processing time is quite long (almost $100$ times more than VJ) thus making it unattractive for realtime implementation on smartphones. However, the best scores are far from satisfactory and better face detectors for AA are needed.\\

\noindent {\bf{Face-based User Verification:}}
Face verification is performed on the UMDAA-02-FD dataset. For each annotated face, $68$ fiducial landmarks are extracted using the Local Deep Descriptor Regression (LDDR) method trained on Imagenet and FDDB datasets \cite{amit_landmarks}. Feature extraction is performed after alignment, centering and cropping.
\begin{figure}[htp!]
\centering
\includegraphics[width=0.35\textwidth]{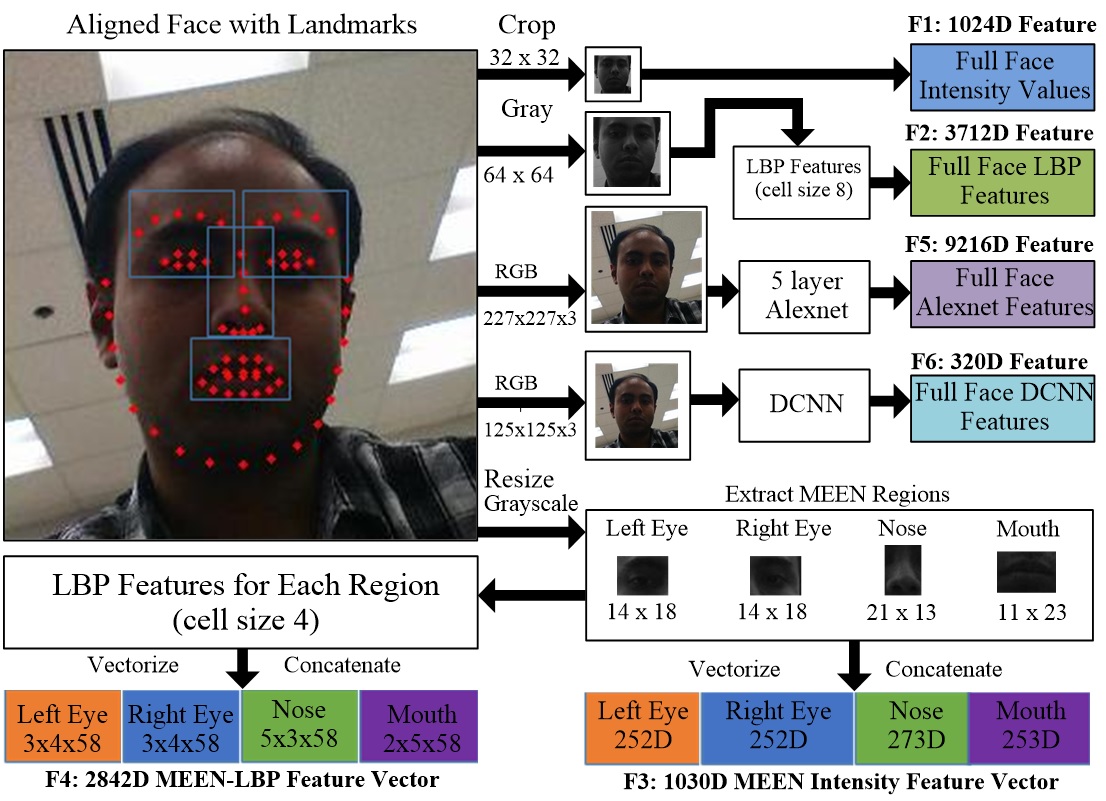}
\vskip -0pt
\caption{Flow diagram for features extraction for face verification.}
\label{FaceFeatExtract}
\vskip -10pt
\end{figure}

\noindent {\bf{Feature Extraction from Faces:}}  Given a face image, pixel intensity, Local Binary Pattern (LBP) \cite{LBP_Ojala_PAMI} and Convolutional Neural Network (CNN) features using the pre-trained Alexnet network \cite{AlexNet_NIPS2012_4824} and the DCNN network \cite{JC_CNNFace} are extracted. In total, $6$ different features are extracted for each face as shown in fig. \ref{FaceFeatExtract}.
\begin{itemize}[noitemsep]
\item $F_1$: Pre-processed faces are converted to grayscale, rescaled ($32\times 32$) and vectorized ($1024$ dimensional vector).

\item $F_2$: From the $64\times 64$ rescaled grayscale image, LBP features of size $8\times 8 \times 58$ ($3712$ dimensional vector) are extracted for a cell size of $8\times 8$ pixels.

\item $F_3$: Bounding boxes of the eyes, nose and mouth are computed from the landmarks with a $5$ pixel margin for each face part from the pre-processed grayscale image. The eyes, nose and mouth bounding boxes are resized to $14 \times 18$, $21 \times 13$ and $11 \times 23$ pixels respectively, then vectorized to a 1030 dimensional MEEN feature\cite{AA_Fathy}.

\item $F_4$: LBP features ($2842$ dimensional) are obtained from each of the resized bounding boxes of MEEN parts ($F_3$) with a cell size of $4 \times 4$ pixels. 

\item $F_5$: The first five convolutional layers of Alexnet are used to extract features of size $6 \times 6 \times 256$ ($9216$ dimensional) from resized color images of faces ($227 \times 227$)

\item $F_6$: Landmarks are input to the DCNN based face verification system  \cite{JC_CNNFace} trained on the CASIA-WebFace dataset \cite{CASIA_WebfaceDset}, which resizes the face to ($125 \times 125 \times 3$) and then outputs a $320$ dimensional feature vector.

\end{itemize}

\begin{figure}[t]
\centering
\includegraphics[width=0.35\textwidth]{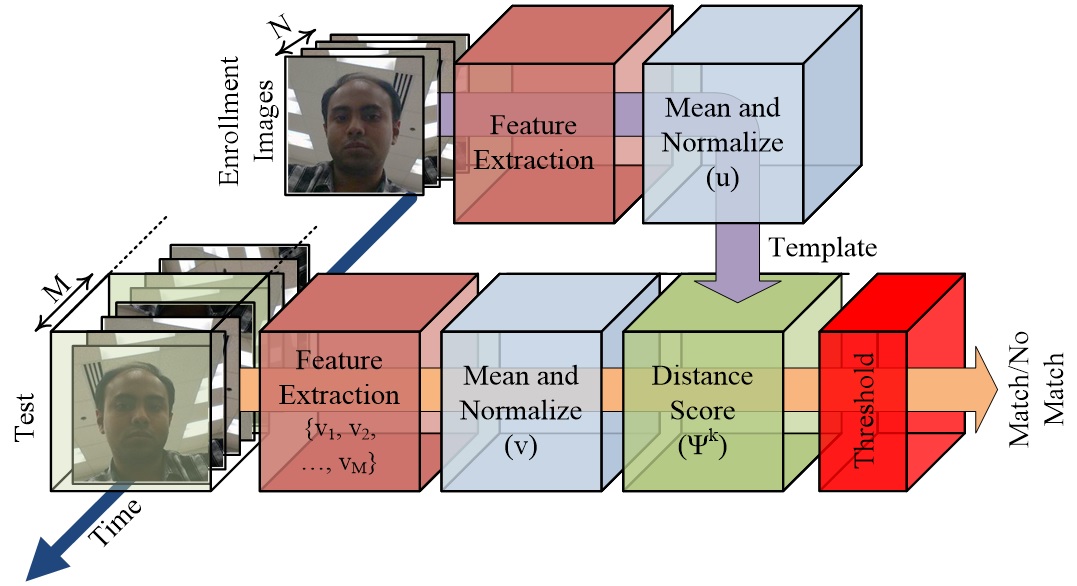}
\vskip -0pt
\caption{Block diagram depicting the face verification protocol.}
\label{ImageVerificationProcess}
\vskip -10pt
\end{figure}

\noindent {\bf{Evaluation Protocol:}}  Six types of feature vectors are considered in this experiment. In the absence of any particular enrollment data, to simulate a practical  AA scenario, the faces are sorted chronologically for each user and the first $N$ faces are considered for enrollment while the rest are used for verification. The mean of the features of the enrollment set of a user followed by $L2$ normalization of the mean vector is stored as his/her template $u$.

Fig. \ref{ImageVerificationProcess} shows a block diagram of the verification process. A reasonable, practical assumption for robust AA is that the user is verified by the last $M$ faces instead of a single one. Therefore features $v_{i}$ ($i=1, 2, \hdots, M$) are extracted from each of the $M$ faces for each location of the moving window, then averaged and $L2$-normalized to form the test vector $v$. The distances between $v$ and $u$ are calculated using four distance measures, namely, Euclidean Distance (EU), Cosine Distance (CosD), Manhattan Distance (MD) and Correlation Distance (CorrD). For the distance measure $\delta^k$ of type $k$ the score is $\Psi^k=\frac{1}{\delta^k}$\cite{AA_Samangouei}.

\begin{figure}[t]
\centering
\includegraphics[width=0.48\textwidth]{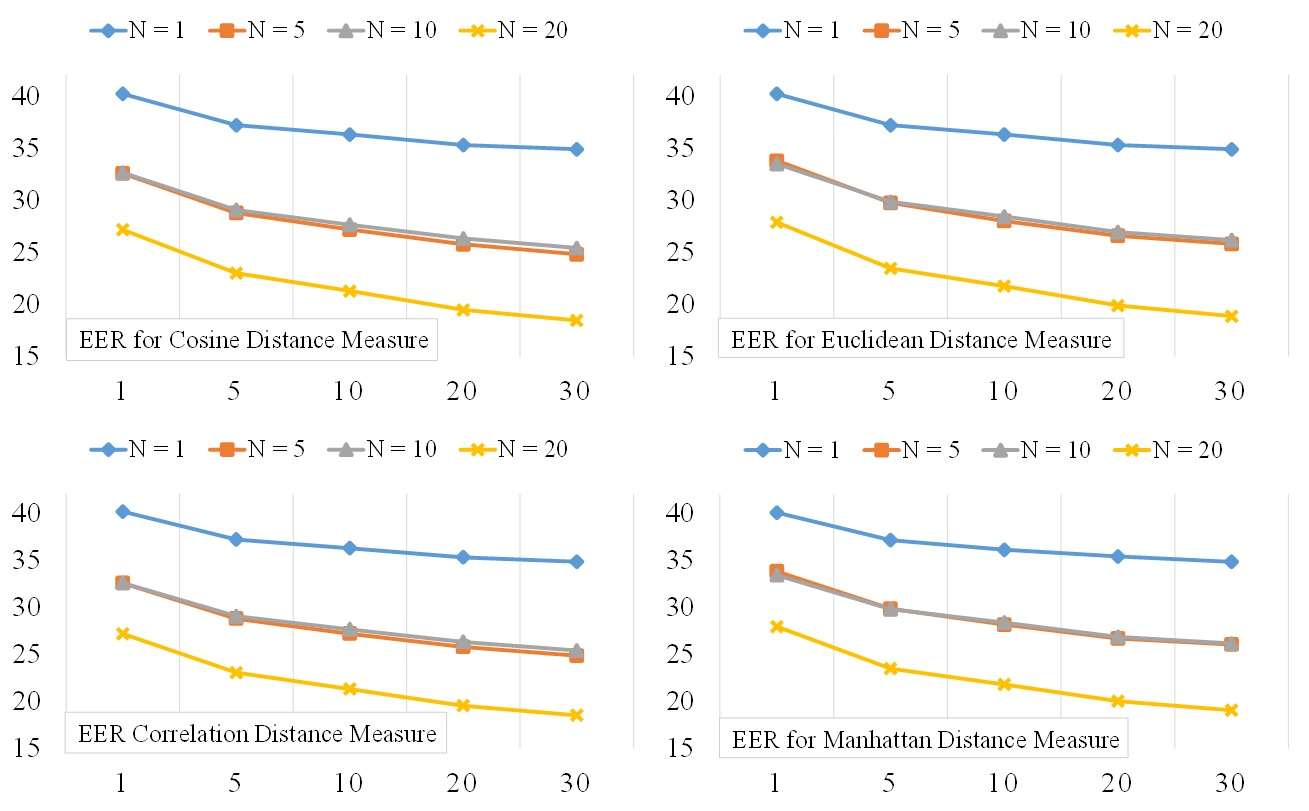}
\vskip -0pt
\caption{EER ($\%$) vs. $M$ for varying $N$ using DCNN features ($F_{6}$) and four different metrics.}
\label{DCNNEERvsDist}
\vskip -10pt
\end{figure}

\noindent {\bf{Experimental Results:}}  In Fig. \ref{DCNNEERvsDist}, the equal error rate (EER) ($\%$) produced by using $F_{6}$ features are plotted for varying $M$ and $N$ values for the four distance measures. It is evident from the plots that the EER decreases with increasing $N$ and $M$ for all the cases. The lowest EER of $18.44\%$ is achieved for $N=20$, $M=30$ using either CorrD or CosD measure. 

\begin{figure}[t]
\centering
\includegraphics[width=0.4\textwidth]{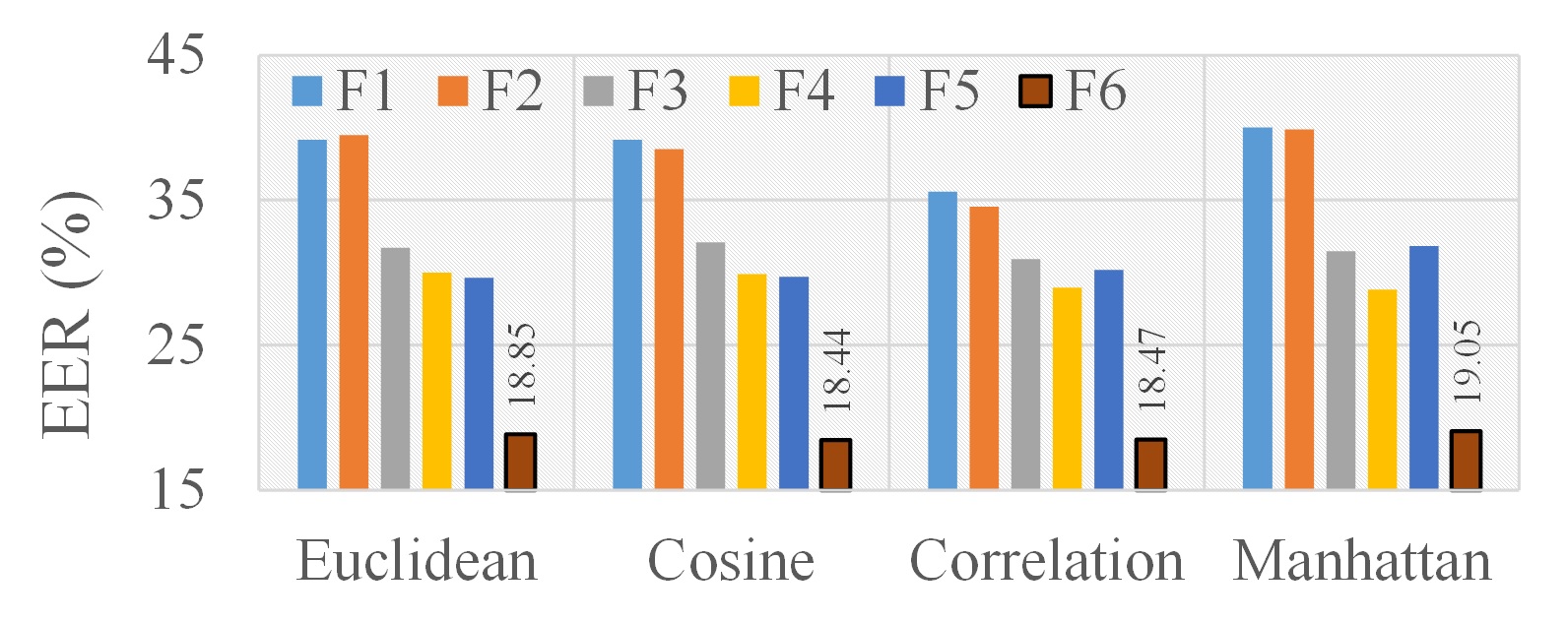}
\vskip -0pt
\caption{EER($\%$) for $6$ feature vectors using four metrics.}
\label{EERvsDist_FT_En20_Sm30}
\vskip -15pt
\end{figure}

Fig. \ref{EERvsDist_FT_En20_Sm30}  shows the EER corresponding to different features and distance measures considering  $N=20$, $M=30$. The DCNN features ($F_{6}$) are found to be the most effective (EER of $18.44\%$ for CosD). Since, for a reliable system the EER is expected to be at least less than $5\%$, this value is not satisfactory at all. The poor performance may be due to the fact that many faces in the dataset are partially visible and therefore alignment using facial landmarks fails badly for these cases. Also, matching the features from a partial face to the features of the same user's full face may result in a large distance measure. Among the other methods, the Alexnet network does not perform much better than the non-CNN features in this scenario as it is not trained particularly for faces. The LBP of MEEN face (EER of $28.83\%$ for MD) gives the best result among non-CNN features. Note that in practice, the CNN feature extraction step is generally much slower than the non-CNN feature extraction methods without the use of a GPU. Thus, more robust yet fast verification methods are needed to produce satisfactory performance on this dataset.

\section{User Identification Using Swipe Dynamics}
\begin{figure}[t]
\centering
\includegraphics[width=0.3\textwidth]{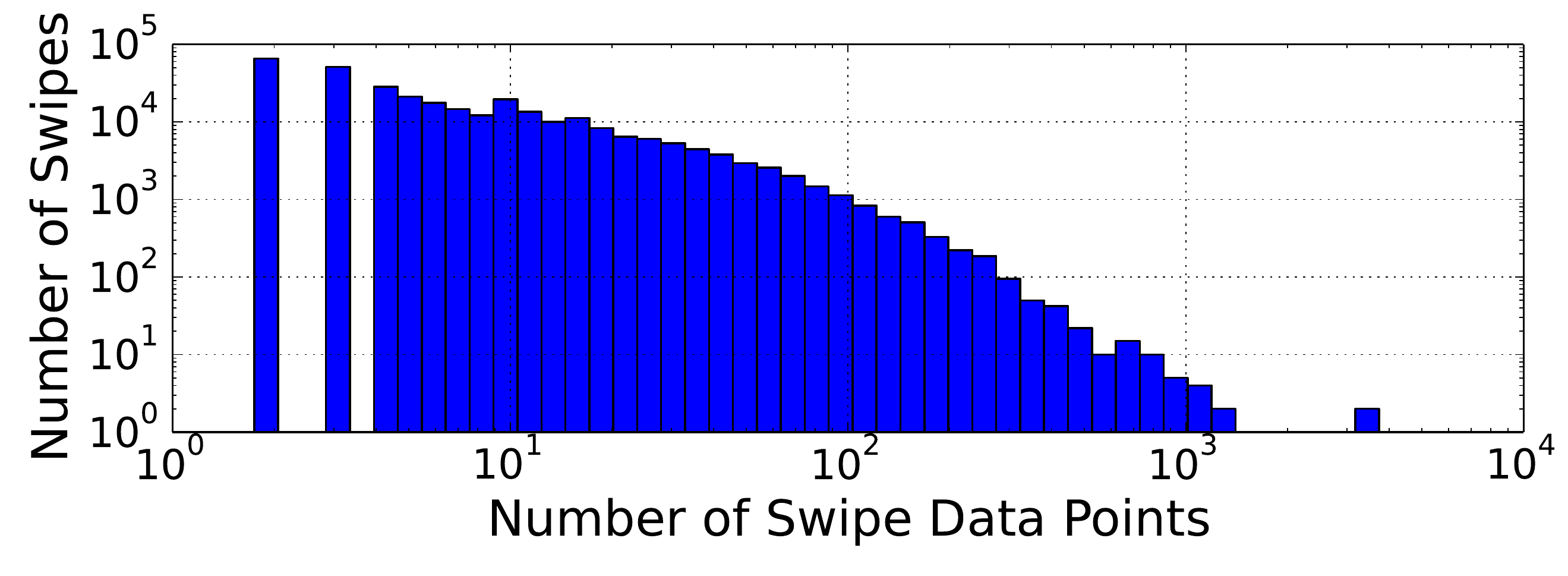}
\vskip -5pt
\caption{Histogram of the number of data points per swipe.}
\label{SwipeLength}
\vskip -5pt
\end{figure}

In this experiment, single finger touch sequences (swipes) on the screen are studied by considering three types of events - finger down, in-touch and finger up.  The length of swipes vary between $1$ to $3637$ touch data points (Fig. \ref{SwipeLength}). For reliable authentication using swipes, longer ones are preferable \cite{Touchalytics}. Hence, swipes with more than four data points are considered for feature extraction. Table \ref{SwipeData} summarizes the swipe dataset, shows that it contains a large number of touch and swipe data per user and therefore can serve as a data set for practical experiments on swipe-based authentication. Since the users were not given any particular task to perform, the touch data in AA-02 is representative of how users interact with the phone through touch.

\begin{table}
\small
\centering
\caption{General Information on Swipe Data}
\begin{tabular}{p{5.8cm} c}
\hline
No. of subjects	                						&$48$\\
\hline
Avg. Session/User with swipe data    	&$\backsim$ $196$\\
\hline
Total taps (finger down-finger up)			& $177417$ \\
\hline
Total swipes (including taps)					& $489723$ \\
\hline
Maximum data points in a swipe					& $3637$ \\
\hline
No. of Swipes/User								&$\backsim$  $10203$ \\
\hline
No. of Swipes/Session							& $\backsim$ $52$ \\
\hline
No. of Swipes ($>4$ data points)	& $\backsim$ $167126$ \\
\hline
No. of Swipes/User ($>4$ data points) &$\backsim$  $3482$ \\
\hline
No. of Swipes/Session ($>4$ data points)							& $\backsim$ $18$ \\
\hline
\end{tabular}
\label{SwipeData}
\vskip -10pt
\end{table}

\noindent {\bf{Feature Extraction:}}
Every swipe $s$ is encoded as a sequence of $4$-tuples
$s_i=(x_i, y_i, p_i, t_i)$
for $i\in {1, \hdots , N_c}$ where  $x_i$, $y_i$ is the location coordinates and  $p_i$ is the pressure applied at time $t_i$. $N_c$ is the number of data points captured during the swipe. From each swipe-action data with $N_c \geq 5$, a $24$-dimensional feature vector, listed in Table \ref{touchFeatures}, is extracted using the method described in \cite{Touchalytics} and \cite{umd_Dataset}. Note, in the UMDAA-02 dataset, the measure of area covered by the finger is not present.

\begin{table}
\small
\centering
\caption{Features Extracted From Each Swipe Event}
\begin{tabular}{p{0.8cm} p{6.2cm}}
\hline
Features 	&Description \\
\hline
\hline
1-2 	&inter-stroke time, stroke duration\\
\hline
3-6 	&start $x$, start $y$, stop $x$, stop $y$\\
\hline
7-8 	&direct end-to-end distance, mean resultant length\\
\hline
9 	&up/down/left/right flag\\
\hline
10-12 	&20\%, 50\%, 80\% -perc. pairwise velocity\\
\hline
13-15 	&20\%, 50\%, 80\%-perc. pairwise acc\\
\hline
16 	&median velocity at last 3 pts\\
\hline
17 	&largest deviation from end-to-end (e-e) line\\
\hline
18-20 	&20\%, 50\%, 80\%-perc. dev. from e-e line\\
\hline
21 	&average direction\\
\hline
22 	&ratio of end-to-end dist and trajectory length\\
\hline
23 	&median acceleration at first 5 points\\
\hline
24 	&mid-stroke pressure\\
\hline
\end{tabular}
\label{touchFeatures}
\vskip -10pt
\end{table}

\noindent {\bf{Experimental Setup and Evaluation:}}
The swipe data for each user (with $N_c \geq 5$) are sorted chronologically and the first  $70\%$ swipes are considered for training-validation while the rest for testing. After extracting the $24$-dimensional feature vector from each swipe, the training feature matrix is normalized to zero mean and unit variance. Then individual binary classifiers are trained for each user following the one-vs-all protocol. The classification methods considered for this experiment are k-nearest neighbor (KNN) \cite{Touchalytics}, Gaussian kernel Support Vector Machine (RBF-SVM)\cite{Touchalytics}, Naive Bayes (NB) \cite{Touch_SerwaddaPW13}, Linear Regression (LR) \cite{Touch_SerwaddaPW13}, Random Tree estimation followed by Linear Regression (RT+LR), Random Forest estimator (RF) \cite{RandomForest_Breiman}, \cite{TouchAA_Feng}, \cite{Touch_SerwaddaPW13} and Gradient Boosting Model (GBM) \cite{GradientBoosting_Friedman00}. The methods are compared based on EER (\%). 

\begin{figure}[t]
\centering
\includegraphics[width=0.45\textwidth]{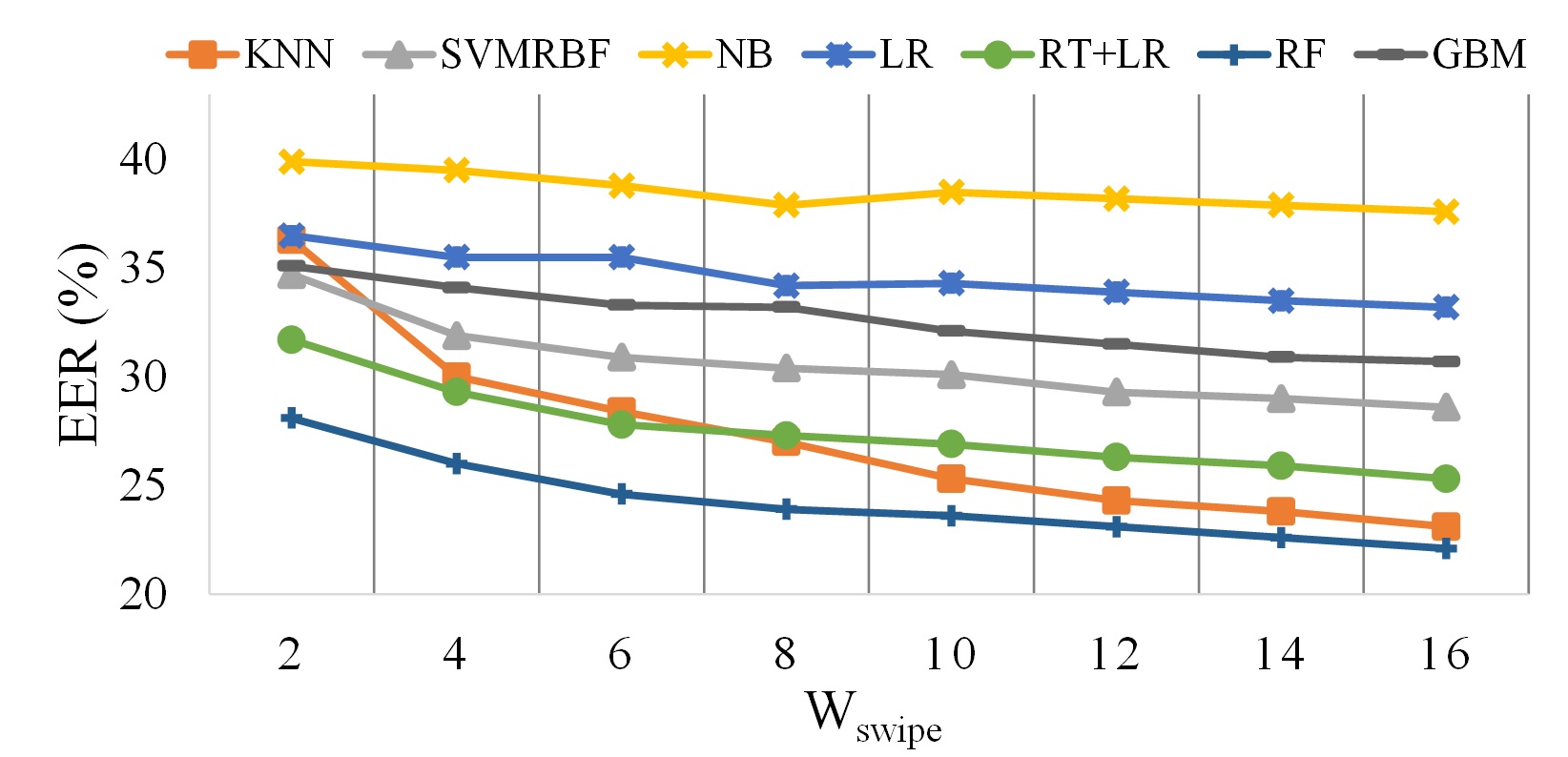}
\vskip -5pt
\caption{EER vs. $W_{swipe}$.}
\label{EERvsNumberOfSwipes}
\vskip -15pt
\end{figure}

As proposed in \cite{Touchalytics}, instead of using a single swipe for authentication, the scores of multiple, consecutive $W_{swipes}$ number of swipes are averaged together for robustness. Since all of the methods return confidence probabilities/scores or distance from separating hyper-plane representing confidence, the score fusion is a simple average of individual scores. For the nearest neighbor-based methods, nine neighbors are considered. The parameters of RBF-SVM are tuned by $10$ fold cross validation on smaller subsets of the original training data. Since the training data is very large, the SVM is trained on a reduced subset, followed by retraining on the hard negative mined error cases. For the ensemble-based methods, the number of estimators is set to $200$ and the maximum tree depth is set to $10$. The EER values obtained using different methods for differnet $W_{swipe}$ values are show in Fig. \ref{EERvsNumberOfSwipes}. The random forest ($RF$) estimation method outperforms all the other methods and can reach an EER of $22.1\%$. However, for practical usage, this EER is not satisfactory and therefore achieving a better performance for this dataset is a new research challenge.

\section{Geo-location Data and Next Place Prediction}
The location service of smartphones return geographical location of the user based on GPS and WiFi network. Excluding the users who kept their location service off, geolocation data, stored only if there is significant change in the location, is obtained from $45$ users (summarized in Table \ref{GeoData}). It is possible to reasonably predict the next location that a person might visit based on prior knowledge on the pattern on one's life. In this section, the next place prediction problem is approached using the geolocation data available in the UMDAA-02 dataset.

\begin{table}
\small
\centering
\caption{General Information on Geo-location Data}
\begin{tabular}{p{6.0cm} c}
\hline
No. of Subjects	                					&$45$\\
\hline
Avg. No. of Sessions/User with Location Data    	&$\backsim$ $186$\\
\hline
Total Number of Location Traces			& $8303813$ \\
\hline
Number of Location Traces Per User			&$\backsim$  $184529$ \\
\hline
Number of Location Traces Per Session			& $\backsim$ $993$ \\\hline
\end{tabular}
\label{GeoData}
\vskip -5pt
\end{table}

\begin{figure}[t]
\centering
\includegraphics[width=0.3\textwidth]{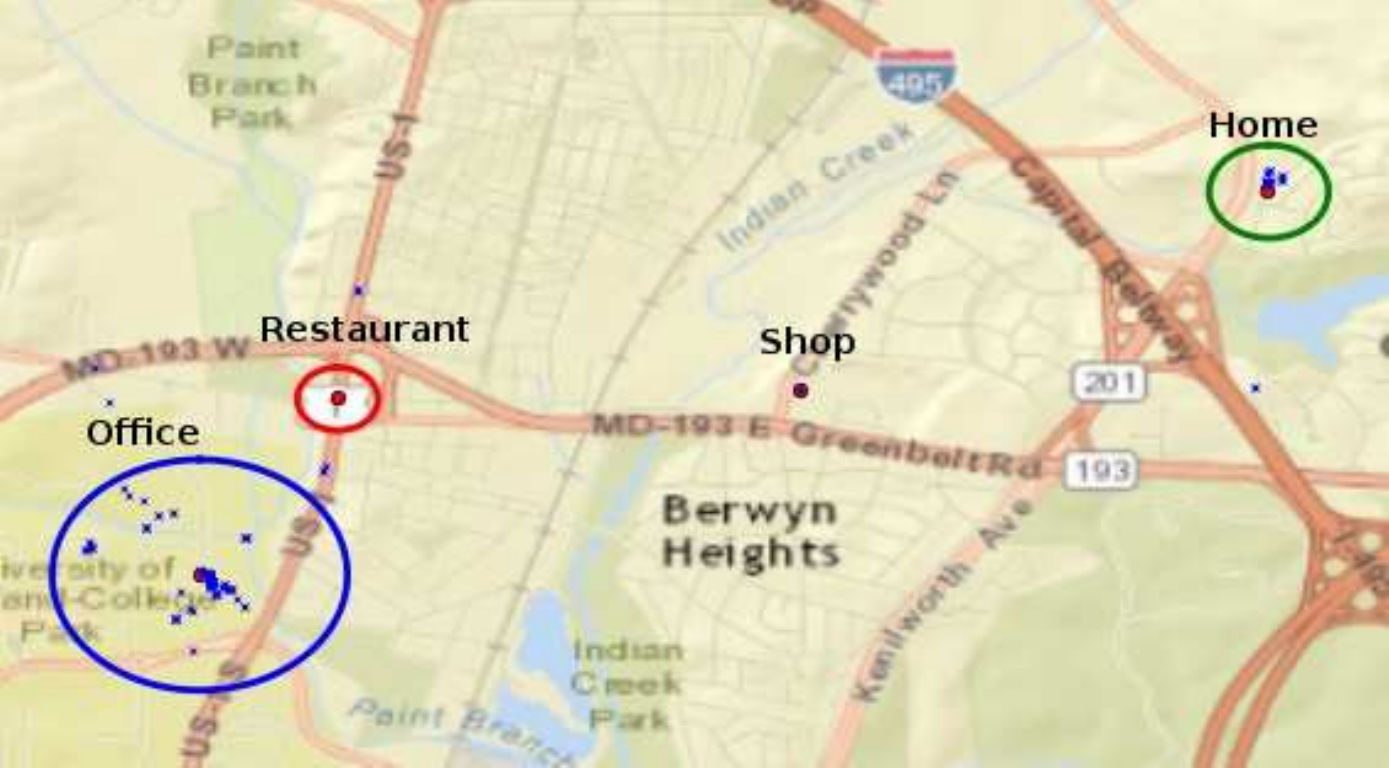}
\vskip 0pt
\caption{Example of Geo-location Data Clustering - Analysis of the clusters reveal states of the user such as 'Home' or 'Work'.}
\label{LocationCLuster}
\vskip -15pt
\end{figure}

\noindent {\bf{State Definition for Mobility Markov Chains:}}  Location histories are first clustered into $N_{i}$ clusters, namely $C^1_i \hdots C^N_i$, for the $i$-th user using the DBSCAN algorithm \cite{DBScanClustering} based on distances between data points. The maximum distance between a point from the center of the cluster in which that point belongs is set to be below a certain value $R$ meters. Such clustering for a student (shown in Fig. \ref{LocationCLuster}) reveals the expected dominant regions that the user would visit - home, university, a certain shop and a restaurant. Two additional clusters, Transit ($Tr$) and Unknown ($Unk$), are also assigned for each user. If the user is traveling, causing location information to change rapidly ($\geq 2ms^{-1}$), then he/she is assigned to $Tr$. The remaining data points are denoted as $Unk$. 

Data points at each cluster are assigned to six different observations based on the day and time information. Weekdays and weekend data points are flagged with $WD$ and $WE$. Also, the whole day is divided into three time zones (TZs) - TZ1 (8:00 am to 4:00 pm), TZ2 (4:00 pm to 10:00 pm) and TZ3 (10:00 pm to 8:00 am). Thus, for the $i$-th user, there are $(N_{i}+2)\times 2\times 3$ possible observation states. However, since the location service only collects data when the phone is unlocked, there are many gaps in the data and it is possible that many of these observation states are absent in the training phase but present in the test data or vice-versa. 

The location service data is utilized for development and evaluation of Mobility Markov Chains (MMC) \cite{MMC_NextPlace}, \cite{MMC_Showme_Gambs} which is a discrete stochastic process model of the mobility behavior of an individual in which the probability of moving to a state depends only on the last visited state and the transition matrix for all probable states.  Thus an MMC is composed of a set of $k$-states $S={s_1, s_2, \hdots, s_k}$, prior probability of entering a state ${p_1, p_2, \hdots, p_k}$ and a set of transitions $t_{i,j}$ where 
$t_{i,j}=Prob(X_n=s_j|X_{n-1}=s_i)$.

\noindent {\bf{Experimental Setup and Evaluation:}}
From the chronological organization of a user's mobility traces, the first $70\%$ are used for training while the rest for testing. Each trace of the training set is tagged with a unique tag identifying the state it belongs to. The prior and transition probabilities of each state are calculated from the chronological traces.  Since, the number of states for a subject depends upon the maximum radius parameter $R$ for the clusters, nearby small clusters get merged into bigger ones with increasing $R$ causing a reduction in the number of states. In the training set, the average number of states per user drops to $35$ from $144$ if the maximum radius is increased to $500$ m from $20$ m. 

\begin{figure}[t]
\centering
\includegraphics[width=0.45\textwidth]{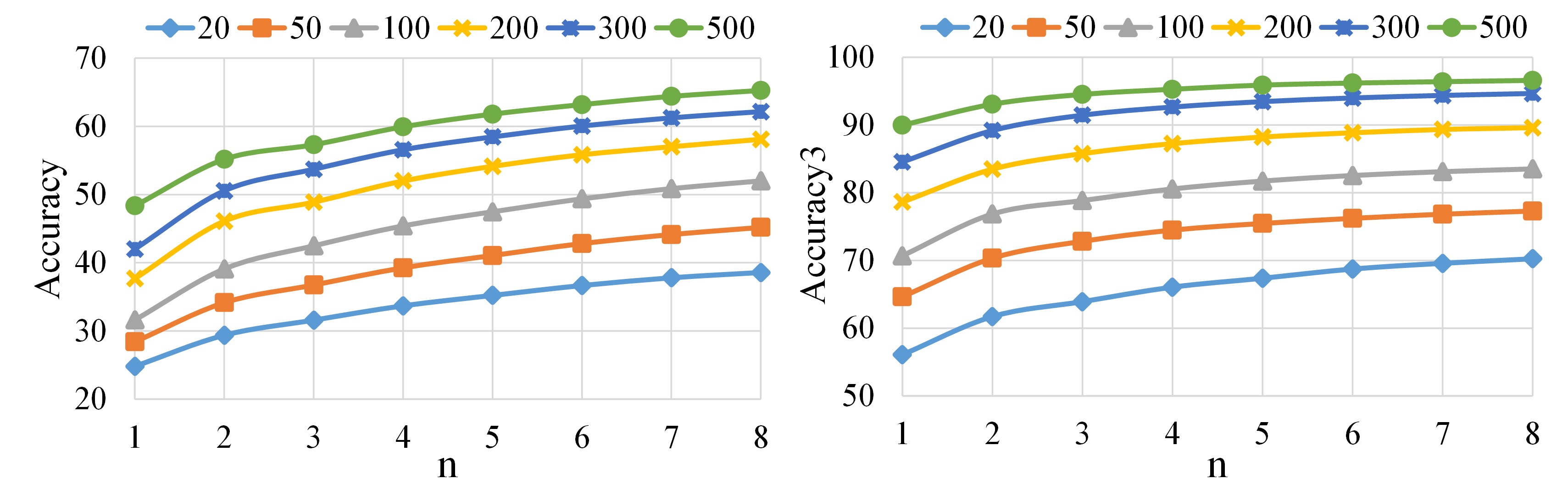}
\vskip -0pt
\caption{Next location prediction $Accuracy$ (left) and $Accuracy3$ (right) measures for increasing number of previous observations for MMC at different $R$.}
\label{AccVsOvs_MaxRadius_Comb}
\vskip -15pt
\end{figure}

MMC-based next location prediction results in terms of $Accuracy$ and $Accuracy3$ (percentage of times the correct next location was among the top $3$ most probable locations) metrics are presented in Fig. \ref{AccVsOvs_MaxRadius_Comb}. The horizontal axis represents the number of previous observations. Considering $n$ previous observations, the MMC algorithm returns probabilities of each state to be the next. Since the day and time zone of the next location are known, states that do not belong to that day and time zone are dropped. The most probable state among the rest of the states is picked as the next predicted location. Fig. \ref{AccVsOvs_MaxRadius_Comb} indicates that knowing more prior states increases the accuracy. The accuracy also increases with increasing maximum radius $R$ (from $20$ meters to $500$ meters) at the cost of localization capability. Between the two measures, $Accuracy3$ is can go much higher ($Accuracy=65.3\%$ and $Accuracy3=96.6\%$ for $R=500$ meters, $n = 8$) indicating the feasibility of location prediction. 

\section{Conclusion}
In this paper, we presented a multi-modal challenge data set for AA problems. Benchmark results for face and touch-based active authentication are provided. Preliminary results for predicting the next location are also given. The UMDAA-02 is the first non-commercial data set on smart phone usage containing data form a wide variety of smart phone sensors. Thus this data set can provide sufficient resources to AA researchers to investigate the efficacy and performance of multi-modal fusion model for a wide variety of modalities in a practical AA scenario. The dataset will be released to the research community in due course.

{\small
\bibliographystyle{ieee}
\bibliography{biblio_PFD}
}

\end{document}